\documentclass[11pt,a4paper]{article}
\usepackage[hyperref]{acl2021}
\usepackage{times}
\usepackage{latexsym}

\usepackage{microtype}

\usepackage{tikz}
\usepackage{pgf}
\usepackage{tikz-dependency}
\usetikzlibrary{spy}
\depstyle{test}{edge below,  label style=trapezium}
\depstyle{cycle}{edge loop,  label style=trapezium}

\usepackage{appendix}
\usepackage[utf8]{inputenc}
\usepackage{epic}
\usepackage{multirow}
\usepackage{graphicx}
\usepackage{amssymb}
\usepackage{tabularx}
\usepackage{tablefootnote}
\usepackage{makecell}

\newcommand{\depedgie}[5][]{
   \begin{scope}
   \depkeys{#1}
   \pgfmathsetmacro{\offa}{#2}
   \pgfmathsetmacro{\offb}{#3}
   \settgtlayer
   \def\source{\wordref{\dt@tgtlayer}{#2}}
   \def\dest{\wordref{\dt@tgtlayer}{#3}}
   \def\depname{#4}
      \def\depname{#5}
   \pgfmathsetlengthmacro{\distance}{abs(\offb - \offa)*\dt@linkstep}
   \groupedge[#1]{\source}{\dest}{\depname}{\distance}
   \end{scope}}

\aclfinalcopy 

\usepackage[acronym, nomain]{glossaries} 
\glsdisablehyper 
\newacronym{ai}{AI}{Artificial Intelligence}
\newacronym{nlp}{NLP}{natural language processing}
\newacronym{cl}{CL}{computational linguistics}
\newacronym{gui}{GUI}{graphical user interface}
\newacronym{ner}{NER}{named entity recognition}
\newacronym{cv}{CV}{computer vision}
\newacronym{dnn}{DNN}{deep neural networks}
\newacronym{r2vq}{R2VQ}{Recipe-to-Video Questions}
\graphicspath{ {./images/} }

\usepackage{hyperref}
\hypersetup{%
        pdftitle={The tikz-dependency package v2.0},
        pdfsubject={A latex package based on tikz to draw dependency graphs},
        pdfauthor={Daniele Pighin},
        pdfkeywords={dependency parses, latex, tikz, pgf},
        colorlinks=true,        
        linkcolor=blue,
        filecolor=blue,
        urlcolor=blue,
        citecolor=blue,
        pdfborder=0 0 0,
}

\title{Designing Multimodal Datasets for NLP Challenges}

 \author{
James Pustejovsky,
Eben Holderness,
Jingxuan Tu, 
Parker Glenn,\\
\textbf{Kyeongmin Rim},
\textbf{Kelley Lynch},
\textbf{Richard Brutti}\\
 Brandeis University
 \\
  \texttt{\{jamesp,egh,jxtu,parkerglenn\}@brandeis.edu}\\ \texttt{\{krim,kelleylynch,richardbrutti\}@brandeis.edu}
}

\date{}

\begin{document}
\maketitle

\begin{abstract}
In this paper, we argue that the design and development of multimodal datasets for \gls{nlp} challenges should be enhanced in two significant respects: to more broadly represent commonsense semantic inferences; and to better reflect the dynamics of actions and events, through a substantive alignment of textual and visual information. We identify challenges and tasks that are reflective of linguistic and cognitive {\it competencies} that humans have when speaking and reasoning, rather than merely the performance of systems on isolated tasks. 
We introduce the distinction between challenge-based tasks and competence-based performance, and describe a diagnostic dataset, \gls{r2vq}, designed for testing competence-based comprehension over a multimodal recipe collection (\url{http://r2vq.org/}).
The corpus contains detailed annotation supporting such inferencing tasks and facilitating a rich set of question families that we use to evaluate \gls{nlp} systems. 

\end{abstract}
\glsresetall

\section{Introduction}
    \vspace*{-2mm}
    
One of the fundamental goals of \gls{ai} has been to create systems that interact with human users fluently and intelligently, by demonstrating inferencing and reasoning capabilities that would be expected of a human partner. To this end, there has been considerable effort recently to create richer datasets in the area of visual and multimodal question answering and inference \cite{wu2018faithful,hu2017learning,johnson2017clevr,antol2015vqa,gao2015you}.
This includes a growing interest in posing larger challenges to end-to-end systems employing architectures with DNNs 
\cite{ribeiro2020accuracy,prabhumoye2020topological}.

These moves are encouraging, yet we argue that there should be a greater focus on linguistic and cognitive {\it competencies} in the field, and not just on question answering skills or ``challenge checklisting''. There are some moves in this direction already \cite{johnson2017clevr}, but there is 
still no generally accepted distinction in current \gls{nlp} between challenge-based tasks and competence-based performance. 
Analogous to human cognitive competencies,     there is both a methodological and modeling advantage to focusing a system's performance on  competence-based learning rather than a narrowly defined task or challenge checklist.
First we define {\it competence-based} learning and testing.

While \citet{chomskyaspects}'s  distinction between
 competence and performance has long been debated in  linguistics, 
the term {\it  competence-based} has been applied
to  a number of different concepts
in both the science of learning and educational communities
\cite{bechtel1999problem,voorhees2001competency,chyung2006building,hsiao2020developing,platanios2019competence}.
The common core to both is a concept capturing a coherent
set of abilities that an individual has in a specific domain.

When designing a challenge (and the dataset supporting it), we need to
distinguish between ``having knowledge of a fact" and ``testing
against the knowledge": having  knowledge of
something is worth little if it cannot be retrieved and
deployed for a specific task. Hence, both tasks and competencies have an 
operational component that is missing from  abstract knowledge of
something. 
However, there is also an important distinction between  task-based and competence-based
behavior. Being able to merely reproduce information does not
presppose understanding of the information. However, when we can apply
our existing knowledge to new situations, we demonstrate a kind of
understanding of how the knowledge (through tasks) is applied. When
viewed over a conceptual domain, this constitutes what we will refer
to as a {\it competence}. A challenge defined in these terms can be
called a {\it competence-based} challenge. 

In Section  \ref{sec:competence}, we describe a model of the queries that can be associated with a competence.  
Section \ref{sec:r2vq}  outlines the requirements on a dataset  reflecting this distinction and describes such a diagnostic dataset, \gls{r2vq}, designed for testing competence-based comprehension over a multimodal recipe collection. 
The corpus contains detailed  annotation supporting meaning representations required of each question family (Section \ref{sec:questions}). 
As a benchmark, we establish baseline evaluation results using several textual and miltimodal statistical leaning methods, and discuss their performance and limitations over the present dataset. 
Section \ref{sec:discussion} discusses our findings and future work.

 \vspace*{-2mm}
\section{Related Work}
\label{sec:related}
    \vspace*{-2mm}
    
\gls{nlp} challenges have helped drive progress in the field for several decades,  from text inference  to question answering \cite{dagan2005pascal,bentivogli2017recognizing,williams2017broad,silva2020xte}. These challenges in part have been framed as specific tasks, and advances on the tasks are largely driven by leaderboards on benchmarks or model comparison on individual datasets. Common benchmarks such as GLUE \citep{glue} and SuperGLUE \citep{superglue} have been used widely. Each benchmark contains several language understanding tasks such as WNLI \citep{WNLI} as an inference task, and WSC \citep{WNLI} as a coreference resolution task. In addition to the benchmark systems, individual tasks have also been developed, such as aNIL \citep{aNLI} and SciTail \citep{SciTail} for text inference; SQuAD 2.0 \citep{squad2}, CoQA \citep{coqa} and ARC \citep{arc} for question answering. 

While all the tasks aim to advance the research towards corresponding \gls{nlp} challenges, whether these reflect human competencies remains a question, especially in recent years with the success of transformers \citep{BERT,xlnet,roberta}. On the GLUE/SuperGLUE leaderboard, many top-ranked models have already performed better than human baseline. Further improvements have come from overfitting to the dataset rather than addressing the challenge \citep{leaderboard}. Current pre-training paradigms may also tune models towards capturing merely statistical patterns, so datasets should be designed to align the models’ ability with human expectations \citep{Linzen2020HowCW}. \citet{Sugawara2020AssessingTB} found that most of the questions from common QA and reading comprehension datasets can be correctly answered by the model without complex reasoning. 

Recent work has been trying to identify and evaluate the tasks that are reflective of  reasoning competencies. \citet{Lyu2020ReasoningAG} proposed reasoning tasks on goal-step and step-step relations from procedural text. \citet{cogs} proposed a semantic parsing dataset that evaluates the human-like compositional generalization of models.  \citet{platanios2019competence} describe a training framework based where training data is presented in a manner consistent with  the model's current competence. 
\citet{ribeiro2020accuracy} designed three test types that can be used to test various linguistic capabilities of \gls{nlp} models. 
Related to the current discussion of multimodal datasets, \citet{youcook2_bb} introduced a method of weakly-supervised video object grounding built on procedural recipe data.


\vspace*{-2mm}
\section{Defining Domain Competence Models}
\label{sec:competence}
    \vspace*{-2mm}
    
This section details the underlying form of the models that characterize a competence. We include the following classes: basic category recognition (ingredients, tools); action and relation recognition (cooking actions, spatial and affordance relations); algebraic operations (counting, comparing sizes, adding and subtracting quantities); temporal reasoning (ordering) of the events denoted by the recipe; spatial inferencing over objects (action location, tool orientation).

\subsection{Objects and Relationships}
\vspace*{-2mm}

`


\begin{figure}
    \centering
    \includegraphics[width=1\linewidth]{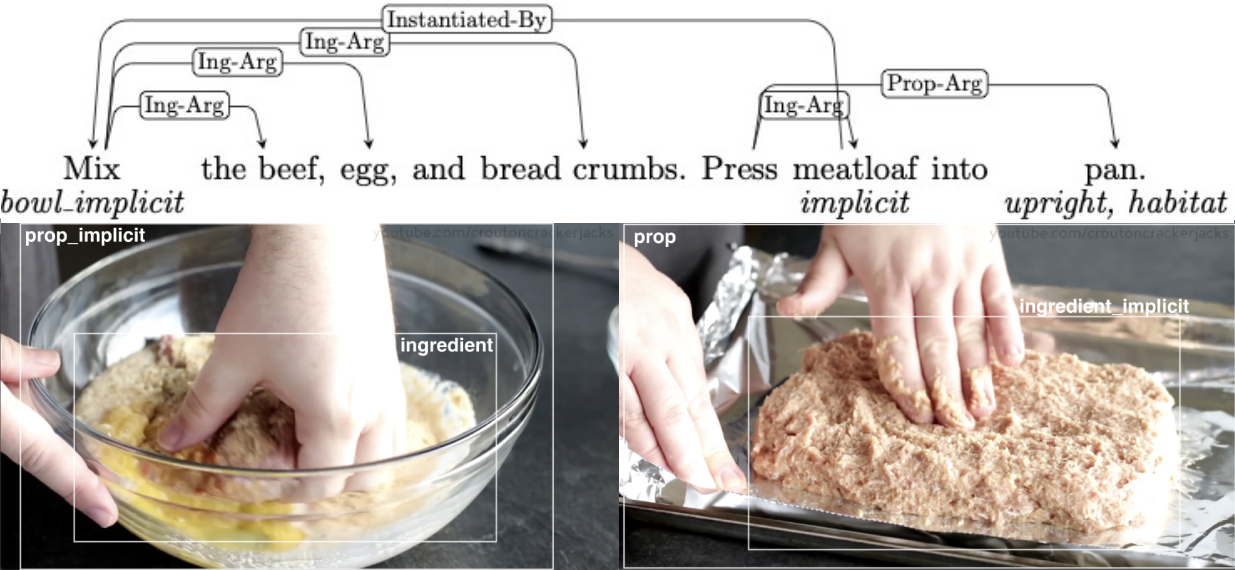}
    \vspace*{-7mm}
    \caption{Example aligned text-video annotation.}
    \label{fig:annotation_example}
    \vspace*{-5mm}
    \end{figure}

\vspace*{-2mm}
\subsection{Recipe Representation}
\vspace*{-2mm}

Traditional cookbook recipes typically have a loose structure; enumerated ingredients followed by arbitrary "steps" that consist of one or more actions. Some ingredients may indicate previous actions (e.g., 1 cup chopped onions). Some actions entail implicit ingredients not otherwise listed (e.g., grease the pan). Ingredients move through the recipe, and are combined and changed in various ways. 
For example, individual ingredients become a mixture, and that mixture is reified as a ``meatloaf'', shown in Figure \ref{fig:annotation_example}. Our representation accounts for these reified objects, and treats them as arguments for subsequent events.

Each recipe action is considered a {\it cooking-event}. We link each {\it cooking-event} to the named {\it ingredients} it requires, as well as named {\it props}. {\it Props} may be {\it tools} (portable implements), {\it habitats} (topoi in which actions take place), or {\it containers} (static receptecales). The semantics of certain verb classes allow for the creation of new objects (examples). We treat these {\it implicit-ingredients} as first class objects, and they can be predicates in {\it cooking-events}.

\vspace*{-2mm}
\subsection{Sentence and Event Ordering}
\vspace*{-2mm}

Sentence ordering is the task of reconstructing coherent text through arranging sentences from the original text. Identifying the order or coherence of text sequences is important in tasks such as multi-document summarization \citep{multisum}, visual storytelling \citep{storytelling} and recipe generation \citep{recipegen}. Understanding procedural text like cooking recipes requires the competency to understand implicit causal effects of mentioned actions on objects, and  object state updates \citep{actiondynamic}. We  evaluate this competency through the ordering of recipe events.


\vspace*{-3mm}
\section{The \gls{r2vq} Corpus}
\label{sec:r2vq}
\vspace*{-2mm}

\subsection{Corpus Characteristics}
\vspace*{-1mm}
\paragraph{Textual component} A collection of recipes sourced from the open-access recipe sites AllRecipes (AR), Epicurious (EP), and Food Network (FN)\footnote{\url{https://www.allrecipes.com/},\\ \url{https://www.epicurious.com/}, \\ \url{https://www.foodnetwork.com/}}. From an initial randomly-selected collection of 18,000 AR, 25,000 EP, and 60,000 FN recipes, we preprocess by removing duplicate recipes for a given dish and ``out-of-distribution'' recipes.\footnote{We use recipe metadata to filter examples with $\leq 3$ or $\geq 10$ steps; with $\leq 3$ or $\geq 20$ ingredients; and examples where
 the average length of each step is $\leq 6$ or $\geq 70$ words.} The remaining ~11,000 examples are clustered based on sentence-level BERT encodings \cite{reimers-2019-sentence-bert} of the titles using K-means clustering with K=100. Finally, for each cluster, we randomly select 20 recipes to create a dataset of 2,000 recipes that cover a wide range of content. 
The 2,000 recipes consist of 503,073 tokens and 8,763 types, with an average of 4.62 steps, 10.8 ingredients, and an average step length of 34.73 tokens per recipe. A total of 51,331 cooking events are annotated, which contain 19,201 explicit ingredients, 16,338 implicit ingredients, 12,316 explicit props, and 11,868 implicit props. 

\vspace*{-2mm}
\paragraph{Aligned video component}
1,101 Youtube videos from the YouCook2 \citep{youcook2} training and validation sets were downloaded and normalized.\footnote{Videos average 5.2 minutes each for a cumulative total of 96 hours. All videos were normalized to 10 fps with a 224 x 224 resolution.} Samples of 32 frames (3.2 seconds) were taken to create video tokens. These were then converted to a joint text-video embedding space using an S3D ConvNet architecture, trained with MIL-NCE loss \citep{s3d, e2e_instructional}. The model, pre-trained on the HowTo100M dataset \citep{howto100}, achieves state-of-the-art results on zero-shot text-to-video retrieval on the YouCook2 dataset. 

Image frames from the YouCook2 dataset were pre-processed at 1fps and retrieved to be aligned and annotated using the timestamps of the most relevant video tokens identified by the S3D MIL-NCE model. Annotators then pair framesets with a span of recipe text containing a \textit{cooking-event}, where a frameset is a 3 image sequence  spanning 3 consecutive seconds in a video. After aligning  them with corresponding framesets, annotators identify \textit{props} and \textit{ingredients} present in each  frame.

\vspace*{-3mm}
\subsection{Annotation and Results} 
\vspace*{-2mm}

Each recipe is annotated at the span-level for cooking-related actions and associated ingredients and props (tools, containers, habitats) as outlined in Section \ref{sec:r2vq}. The  annotation methodology is described in the Appendix. 

 








  \begin{table}
        \centering
        \resizebox{\linewidth}{!}{
        \begin{tabular}{ll}
        \hline
             \textsc{Order} & \textsc{Recipe Event Sentences} \\
              \hline
            \makecell{4} & \makecell[l]{\textit{Pour \textbf{olive oil} over the garlic and season with salt.}}\\
            \hline
            \makecell{8} & \makecell[l]{\textit{Reserve remaining \textbf{olive oil}.}}\\
            \hline
            \makecell{14} & \makecell[l]{\textit{Toss green beans, roasted garlic and reserved \textbf{olive oil}, toasted} \\ \textit{almonds, and 1 tablespoon olive oil together in a large bowl; ...}}\\
            \hline

        \end{tabular}}    
        \vspace*{-3mm}
        
        \caption{Illustration of event ordering. All sentences that mention the ingredient ``olive oil'' from a recipe are annotated for event ordering task. \textsc{Order} shows the position of the sentence in the original recipe.\protect\footnotemark}
        \label{event-order}
        \vspace*{-4mm}
    \end{table}

\footnotetext{\url{https://allrecipes.com/recipe/230750/}}

For event ordering labeling, first we use string and word embedding matching between ingredient list and instructions to find subsets of sentences that explicitly mention the same ingredient. We then use the annotated ingredient coreference to find implicit mentions. Table \ref{event-order} shows a sample set of recipe event sentences. These sentences are discontinuous in the original recipe, requiring the competency to understand more about the action/object dynamics to generate correct ordering. For example, the event {\it reserve remaining olive oil} should happen naturally after most olive oil has been consumed (action {\it pour}); the object {\it reserved olive oil} can only be generated after t he action {\it reserve}. 

We analyzed the  model's performances on recipe event ordering. The result suggests that event ordering is  more challenging and the model can be further improved on learning action and object dynamics in order to capture the true order of events. Details are provided in the Appendix.

\vspace*{-3mm}
\section{Challenge Tasks on \gls{r2vq}}
\label{sec:questions}
\vspace*{-2mm}

Given the intuition that textual and visual information  mutually inform each other for semantic reasoning, we include the following challenge tasks, designed to involve aligned text-video objects. 

\vspace*{-2mm}
\subsection{Question Families}
\vspace*{-2mm}

We adopt the concept of ``question families'' as outlined in the CLEVR dataset \cite{johnson2017clevr}. While some question families naturally transfer over from the VQA domain (e.g., integer comparison, counting), other concepts such as ellipsis and object lifespan must be employed to cover the full extent of competency within procedural texts.
Based on the competencies detailed below, we will be providing autogenerated queries alongside the annotated recipes.\footnote{We adopt a strategy similar to that used in \citet{johnson2017clevr}.}

\vspace*{-2mm}
\paragraph{Cardinality}
This group covers   concepts of integer comparison and counting. Example questions include \emph{``How many bowls are used?''} and \emph{``Are there more baking pans or bowls?''}.

\vspace*{-2mm}
\paragraph{Ellipsis}
This group deals with identifying arguments that are omitted from a text, but can be understood from context. For example, a verb phrase \emph{``Add the eggs''} may not have the explicit prepositional phrase \emph{``...in the bowl''} in a typical recipe.

\vspace*{-3mm}
\paragraph{Implicit Argument Identification}
This question family covers both implicit tools and habitats introduced in the text. This is distinct from ellipsis, as these questions can not be solved merely through contextual clues. Instead, they require  competence; applying knowledge of an action and its requirements to a novel situation. Object/action pairs may select for a specific, but unmentioned, habitat. \emph{``Drain the pasta''} will select the habitat of \emph{``(over) the sink''}. Others may require a portable implement (\textit{tool}), such as \emph{``Beat the eggs''} requiring a utensil, most commonly being a \emph{``fork''}. 

\vspace*{-2mm}
\paragraph{Object Lifespan/Coreference}
Any system displaying full competency of text should be expected to track the dynamic state of an object when acted upon by another agent. This set of questions deals with state-change concepts such as a \emph{``batter''} becoming a \emph{``cake''} only when acted upon by the event \emph{``bake''}. An example of an object lifespan question may be \emph{``In step 3, where is $<$INGREDIENT$>$?''}.

\vspace*{-2mm}
\paragraph{Object Orientation}
In procedural texts, there is an explicit relationship between the orthographic and physical domains. A fully competent system should be able to express an understanding of this relationship, answering questions such as \emph{``How is the \textit{bowl} oriented during the action ``Pour the batter into the baking pan''?''}

\vspace*{-3mm}
\paragraph{Event Ordering} This set of questions considers the order of a set of specific event sentences from the procedural text. Examples include: \emph{``What is the order for processing green beans in this recipe?''} or \emph{``which comes first - cooking or draining bacon?''}. A competent system should be able to show a comprehensive understanding of object lifespan and action dynamics to answer such questions.

\vspace*{-3mm}
\section{Conclusion and Future Work}
\label{sec:discussion}
\glsreset{r2vq}

\vspace*{-3mm}

In this paper, we present a {\it competence-based} evaluation  strategy  as a new approach for designing NLP challenges,  in order to better characterize the underlying operational knowledge that a system has for a  conceptual domain, rather than focusing on individual tasks. To support such an evaluation, we have  created a multimodal corpus, \gls{r2vq}, annotated for generation of competence-based question families for evaluation. We are currently adding both single and multi-sentence AMR annotations, providing an additional training corpus for both text-only  and multimodal AMR parsing algorithms.

\section{Corpus Annotation Requirements}
\label{appendix:requirements}
Here we describe the requirements on the R2VQ corpus annotation (\url{http://r2vq.org/}). 
Each  recipe  is  annotated  at  the  span-level  for cooking-related actions and associated the ingredients and props (tools, containers, habitats). The ingredients can be either labeled as explicit (those listed in the ingredients section of the recipe) or implicit (the intermediate outputs of applying a cooking action to a set of explicit ingredients). Additionally, we include a variety of attributes that allow for iterative state tracking over the recipe text: each cooking event is directionally associated with the cooking action preceding it, allowing for a trace of ingredients and props as they are modified by each action \cite{malmaud2014cooking}, as well as coreference grounding for implicit ingredients (e.g. the implicit ingredient \textit{marinade} is associated with the cooking event \textit{combine(vinegar,soy\_sauce,oil)}). Props are also annotated for orientation, which provides additional contextual information for downstream visualization and semantic reasoning tasks. 

For each recipe we consider the natural order of cooking steps and the order of sentences from each step as the correct order in the annotation. We define event ordering as the task of identifying the order of sentences that mentioned events of the same ingredient. 

In the example sentence, \emph{In a large pot, mix the white vinegar, sugar, water, cinnamon, salt, and cloves together}, each of the comestibles receive its own \emph{Ingredient} label. All are individually linked to the \emph{Cooking-Event} ``mix'' with a \emph{Cooking-Arg}. ``Mix'' is also linked to the \emph{prop} ``pot'' with a \emph{Prop-Arg}. Annotators further describe the ``pot'' as the \emph{habitat} for the ``mix'' action, and that its spatial \emph{orientation} is in the default, upright position. In the subsequent sentence,  \emph{bring the mixture to a boil}, ``mixture'' is tagged as an \emph{implicit-ingredient} and linked to the new \emph{cooking-event} ``boil''. Additionally, ``mixture'' is related to the previous \emph{cooking-event} ``mix'' with an \emph{Instantiated-by} link. This proves useful for co-reference tasks, as well as tracking the output of the reification contained in \emph{cooking-events}.

\section{Detailing Existing Datasets}
\label{appendix:existing}
When sourcing the video data, ~1,000 videos in the YouCook2 training and validation splits were used. A related dataset released in 2018, the YouCook2-BoundingBox \citep{youcook2_bb} dataset is a step forward in the right direction of visual-semantic grounding in a multimodal dataset, but still lacks the ability to display the full notion of competency outlined in our paper. The dataset contains 15,000 video-description pairs, annotated with the bounding boxes of the 67 most frequent objects. Of the 60,663 seconds (~17 hours) of video data annotated with bounding boxes in the validation set of YouCook2-BB, 150,647 objects were annotated. Of the 150,647 annotated objects, 26,094 of those objects are occluded from view. 

It is important to note that objects are only annotated when explicitly mentioned in a given text description. As a result, the competency-based inference that the action ``Beat the eggs'' requires a ``fork'' is not accounted for in the YouCook2-BoundingBox dataset. Our alignment of framesets with contextualized  \textit{cooking-actions} and annotation of all \textit{props} in the frames account for these sort of  competency-based inferences.

\section{Image Frame Post-Processing}
\label{appendix:images}

In an attempt to filter out discontinuous frame sets (e.g., where the camera abruptly zooms out to show the chef), we run a simple pre-trained face detection model over the frames returned by the S3D MIL-NCE. Any frameset with an instance of a face appearing or disappearing across consecutive frames was discarded. This heuristic worked well across the YouCook2 dataset as an indicator of an undesired scene change, and improved continuity between the video data aligned with text.

\begin{figure}[h!]
  \centering
  \includegraphics[width=1\columnwidth]{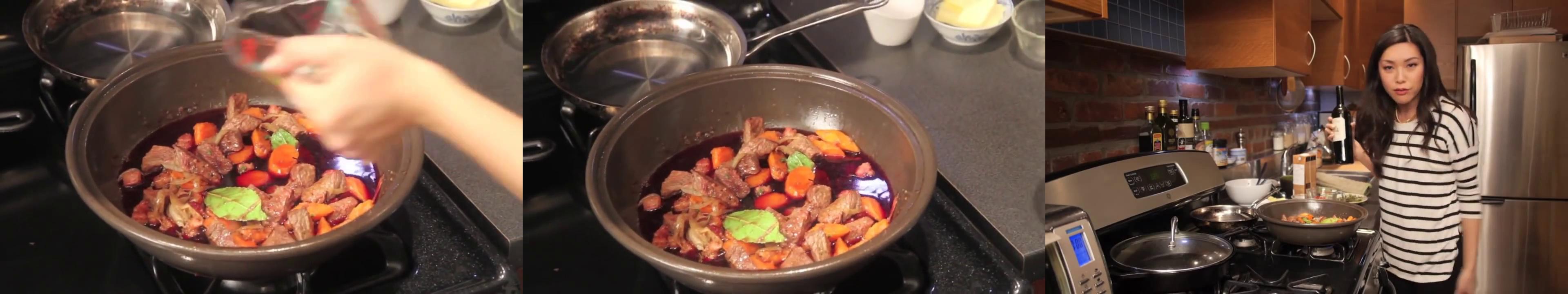}
  \caption{Example of a frameset discarded after post-processing, as a face appears between frames 2 and 3.}
  \label{post_process_example}
\end{figure}

\section{Vector Quantization of Video Tokens}
Using a method of vector quantization similar to that of \citet{videobert}, we cluster the global pooled video embeddings from the S3D ConvNet with hierarchical agglomerative clustering. We set $K=2100$, using the token-to-cluster ratio seen in \citet{videobert} as reference and verifying with a simple elbow method analysis.   

\begin{figure}[h!]
  \centering
  \includegraphics[width=1\columnwidth]{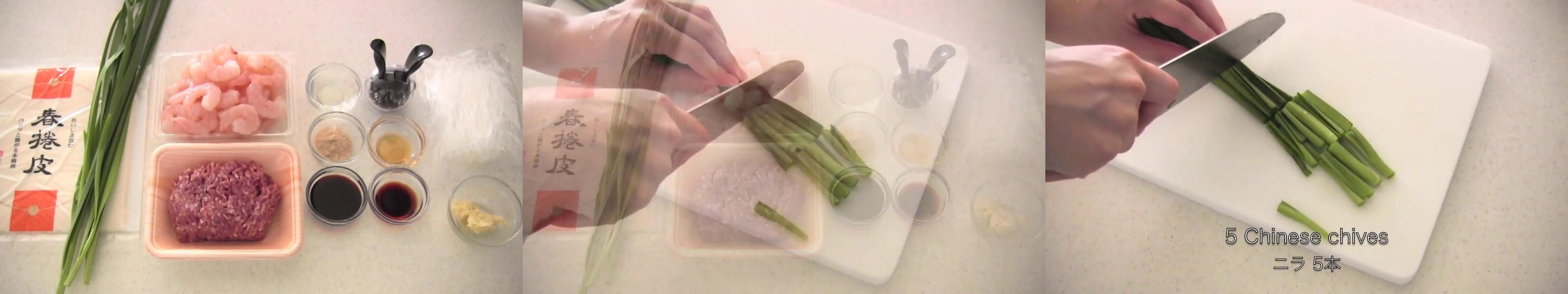}
  \caption{Frameset returned for ``Chop the green onions'' before clustering.}
  \label{original_example}
\end{figure}

\begin{figure}[h!]
  \centering
  \includegraphics[width=1\columnwidth]{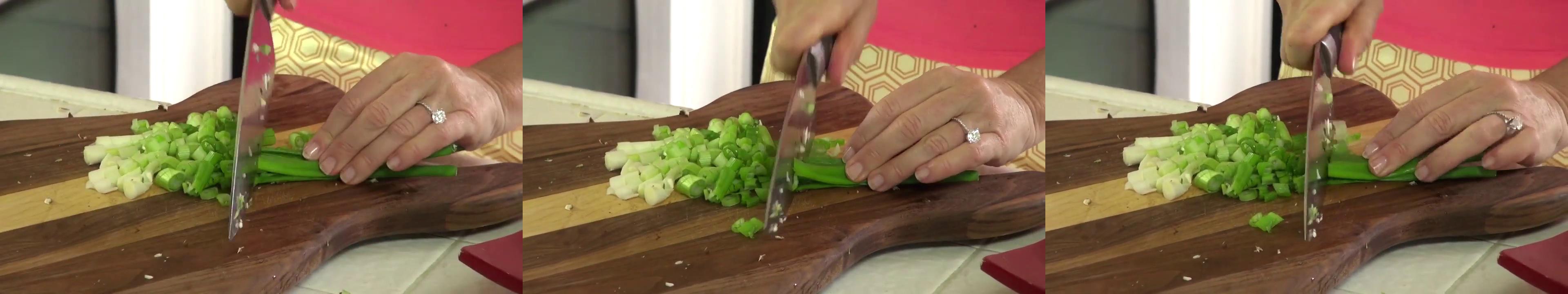}
  \caption{Visual centroid for ``Chop the green onions''.}
  \label{centroid_example}
\end{figure}

After assigning each of the 98,863 3.2 second video tokens to a cluster, token centroids are found. Depicted in Figures \ref{original_example} and \ref{centroid_example}, these token centroids can be useful in filtering out noise seen across framesets depicting identical cooking actions. The goal is to assign the most optimal, ``semantically salient'' frames to each cooking action for further verification by human annotators. In addition to the most relevant framesets identified by the S3D MIL-NCE model, we will be providing annotator's with the frameset representative of the highest-rated video token's visual centroid during the text-video pairing stage of annotation.

\section{Event Ordering Evaluation}

All recipes from R2VQ have more than two sentences, so for experiment when $2 = N$, we randomly sample consecutive sentences pairs from recipes and use those as documents of two sentences. We do not consider the setting when $11 \leqslant N$, becasue there is only one set of \textsc{Event} exceeds that length.

\label{appendix:eventorder}
  \begin{table}
        \centering
        \resizebox{\linewidth}{!}{
        \begin{tabular}{l|lrrr}
        \hline
             & & \textsc{PMR} & \textsc{Acc} & \textsc{Tau}\\
              \hline
            $2 = N$ & \textsc{Sent.} & \textbf{92.06}& \textbf{92.06} & \textbf{0.84}\\
            & \textsc{Event} & 79.52& 79.52 & 0.59\\
            \hline
          $3 \leqslant N < 6$ & \textsc{Sent.} & \textbf{44.86}& \textbf{61.89} & \textbf{0.66}\\
            & \textsc{Event} & 38.73& 54.74 & 0.47\\
            \hline
            $6 \leqslant N < 11$ & \textsc{Sent.} & \textbf{8.41}& \textbf{39.30} & \textbf{0.61}\\
            & \textsc{Event} & 0.00& 25.53 & 0.29\\
            \hline
        \end{tabular}}
        \caption{Sentence (\textsc{Sent.}) and event (\textsc{Event}) ordering results on documents with different number of sentences.\protect\footnotemark}
        \label{tsor-res}
    \end{table}

We analyze the performance on recipe event ordering of the BERT topological sort model from \citep{tsort}. We adopt the commonly used Accuracy (Acc), Perfect Match Ratio (PMR) and Kendall’s tau (Tau) as the metrics to evaluate the performance.  Table \ref{tsor-res} shows the results for all sentences (\textsc{Sent.}) and event sentences (\textsc{Event}) from recipes. We make a head-to-head comparasion between \textsc{Sent.} and \textsc{Event} under three settings where the document length is different. Under each setting, \textsc{Event} constantly performs worse than \textsc{Sent.} on all three metrics, suggesting event ordering is much more challenging and model can be further improved on learning action and object dynamics in order to capture the true order of events. One extreme case is when $6 \leqslant N < 11$, the model is not able to make a perfect match prediciton (\textsc{PMR} is 0) for any set of recipe event sentences. 

\clearpage
\bibliographystyle{acl_natbib}
\bibliography{acl2021} 

\end{document}